\documentclass{article}
\usepackage{spconf,amsmath,graphicx,subfigure}

\title{MixModule: Mixed CNN Kernel Module for Medical Image Segmentation}
\name{Henry H. Yu$^{\star \dagger}$ \qquad Xue Feng$^{\star}$ \qquad Ziwen Wang$^{\dagger}$ \qquad Hao Sun$^{\ddagger}$}
\address{$^{\star \dagger}$ Tsinghua University, afishyay@gmail.com \\
     $^{\star}$ University of San Francisco, xfeng23@dons.usfca.edu \\
     $^{\dagger}$ Boston University, asoapqaq@bu.edu\\
     $^{\ddagger}$ University of Southern California, sh1005536292@163.com}

\begin{document}
%
\maketitle
\begin{abstract}
Convolutional neural networks (CNNs) have been successfully applied to medical image classification, segmentation, and related tasks. Among the many CNNs architectures, U-Net and its improved versions based are widely used and achieve state-of-the-art performance these years. These improved architectures focus on structural improvements and the size of the convolution kernel is generally fixed. In this paper, we propose a module that combines the benefits of multiple kernel sizes and we apply the proposed module to U-Net and its variants. We test our module on three segmentation benchmark datasets and experimental results show significant improvement.

\end{abstract}
\begin{keywords}
Semantic segmentation, U-Net, R2U-Net, Attention U-Net, Mixed Kernels
\end{keywords}
\section{Introduction}
\label{sec:intro}

In recent years, the rapid development of deep learning technology(DL) represented by CNNs has greatly promoted the advancement of computer vision research fields such as classification, detection, segmentation and tracking. Many excellent CNN models like AlexNet\cite{krizhevsky2012imagenet}, VGG\cite{simonyan2014very}, GoogleNet\cite{szegedy2015going}, ResNet\cite{he2016deep}, etc. have been proposed and achieved good results since 2012. In addition, researchers have also established datasets such as ImageNet\cite{deng2009imagenet}, COCO\cite{lin2014microsoft}, etc. which have also greatly promoted the development of related research. Initialization using pre-trained model parameters on these datasets also greatly increases the efficiency of the study. In view of the great success of DL in the field of computer vision, researchers have applied it to medical images such as Computer Tomography (CT), ultrasound, X-ray and Magnetic Resonance Imaging (MRI), in which the development of automatic segmentation technology has effectively reduced the time and cost of manual labeling and it is also the objects of this study. A big difference between medical images and natural scene images is that medical images are difficult to obtain. The scarcity of data leads to the inability to train deeper neural networks, and the domain gap of medical images and natural images also leads to bad performance on models pre-trained on ImageNet, COCO and other natural scene datasets. In this case, U-Net and different improved versions based on it are proposed and achieve good segmentation performances with relatively few datasets. These variants focus on improvements in network architecture, such as the integration of recurrent neural networks into U-Net, etc. In this paper, we study the effect of the convolution kernel size to the performance of the model and propose a new module named MixModule. We expect different sizes of convolution kernels to capture different levels of information since they have different receptive fields and the fusion of these information plays an extremely important role in improving network performance.

\section{Related Work}
\label{sec:rel}

Semantic segmentation which classifies each pixel in the image individually is an important research area in computer vision. Before the advent of DL revolution, traditional methods mostly rely on manual extraction of features to predict the category of each pixel. Even in the early days of DL, researchers mainly use patch-wise training\cite{ciresan2012deep}\cite{farabet2012learning}\cite{pinheiro2014recurrent} which classifies pixels by using an image block around the pixel as input to feed into the CNN. This method is not only inefficient since the content of adjacent pixel blocks is basically repeated but also limits the sensing area due to the pixel block size, so it is difficult to achieve good results. The FCN method proposed by Long et al.\cite{long2015fully} that uses a fully convolutional network structure and applies fully convolutional training has completely changed the situation and became the basis of subsequent research. The deeplab series of studies\cite{chen2014semantic}\cite{chen2017deeplab}\cite{chen2017rethinking}\cite{chen2018encoder} based on FCN propose Atrous convolution and other operations to further improve the accuracy of semantic segmentation. On the basis of semantic segmentation, instance segmentation study is developed which not only predicts the pixel class, but also predicts the class individuals to which the pixel belongs. Mask RCNN\cite{he2017mask} method based on Faster RCNN\cite{ren2015faster} and PANet\cite{liu2018path} method based on FPN\cite{lin2017feature} have achieved the state-of-the-art performance on the instance segmentation task. Although aforementioned methods have achieved impressive results, they are based on pre-trained features on public datasets such as ImageNet\cite{deng2009imagenet}. Due to the domain gap between medical images and natural scene images and the scarcity of medical images, the above methods cannot be well transferred to the medical image segmentation task. In response to the characteristics of medical images, Olaf et al. propose U-Net\cite{ronneberger2015u}, which achieves competitive performance using a relatively small number of medical images. On the basis of U-Net, researchers have successively proposed R2U-Net\cite{alom2018recurrent}, Attention U-Net\cite{oktay2018attention}, etc., which further promote the development of medical image segmentation research. This paper proposes MixModule from the perspective of convolution kernel size and demonstrates its contribution to the performance of U-Net and its variants.

\section{Proposed Method}
\subsection{MixModule}
\label{ssec:mixmodule}

MixModule contains multiple sizes of convolution kernels to capture different ranges of semantic information which is crucial for medical images that emphasize the details of the underlying image. Let $W_i^{(k,k,c,m)}$ denotes the $i$th convolutional kernel whose kernel size is $k\times k$, input channel size is $c$, filter number is $m$. Let $X^{(h,w,c)}$ denotes the input tensor with height $h$, width $w$ and $c$ channels. Let $Y_i^{(h,w,m)}$ denotes the $i$th output tensor using $W_i^{(k,k,c,m)}$ which is calculated as \eqref{con:Yi} and the output tensor $Y^{(h,w,m\times n)}$ is obtained by concatenating $Y_i^{(h,w,m)}$\eqref{con:Y} where $n$ is the total number of kernels used.

\begin{small}
\begin{equation}
Y_{x,y,z}=\sum_{-\frac{k}{2}\leq i \leq\frac{k}{2},-\frac{k}{2}\leq j \leq\frac{k}{2}}{X_{x+i,y+j,z}\cdot W_{i,j,z}}\label{con:Yi}
\end{equation}
\end{small}
\begin{small}
\begin{equation}
Y^{(h,w,m\times n)}=Concat(Y_1^{(h,w,m)}, ..., Y_n^{(h,w,m)})\label{con:Y}
\end{equation}
\end{small}

\begin{figure}[htb]
\centering
\subfigure[Conv block]{
\label{fig1a} 
\includegraphics[width=2.0cm]{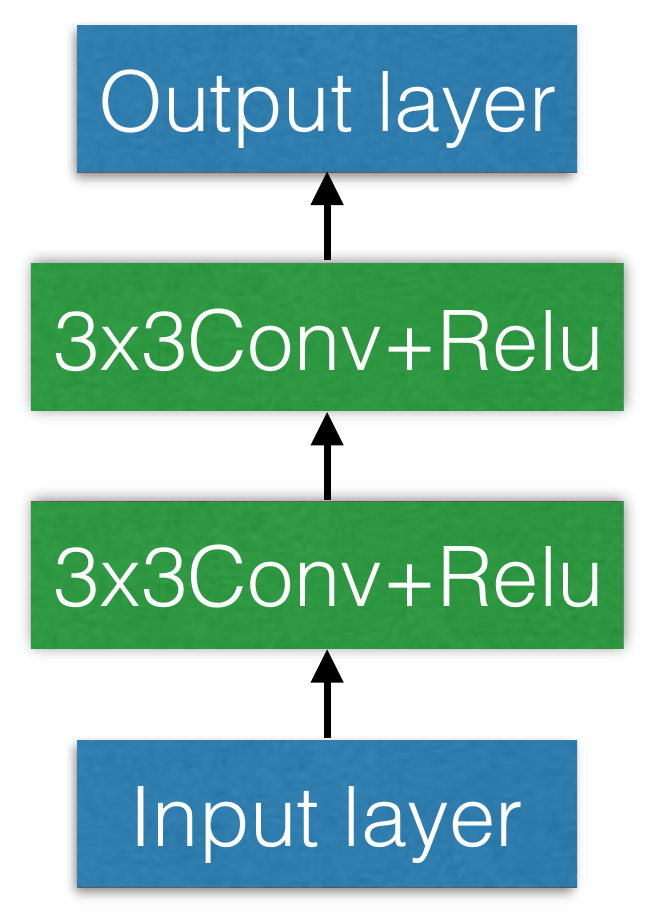}}
\hspace{1.0cm}
\subfigure[Recurrent Conv block]{
\label{fig1b} 
\includegraphics[width=3.0cm]{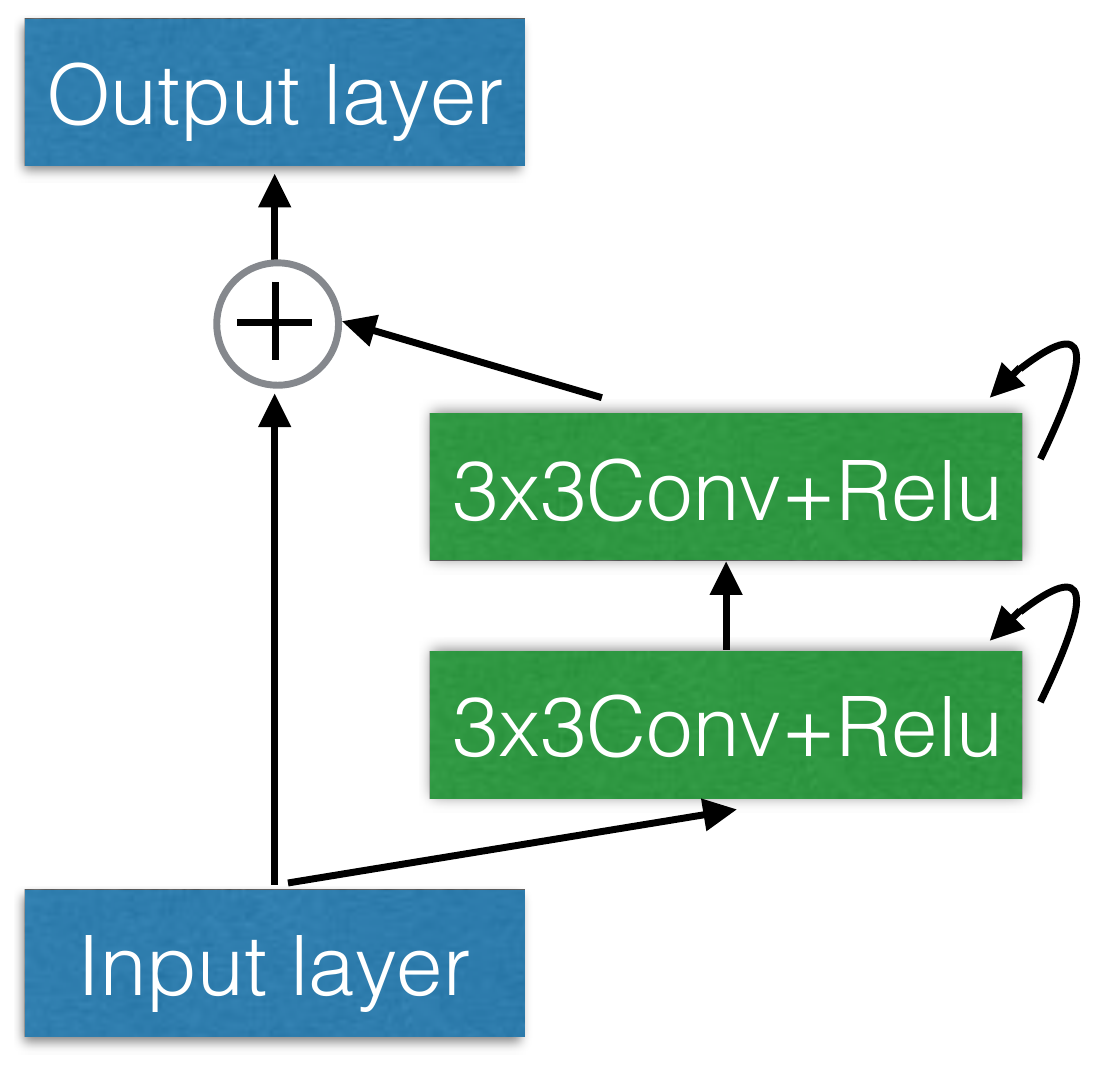}}
\caption{Basic block in U-Net and its variants}
\label{fig1} 
\end{figure}

\begin{figure}[htb]
\centering
\subfigure[mix conv block]{
\label{fig2a} 
\includegraphics[width=3.0cm]{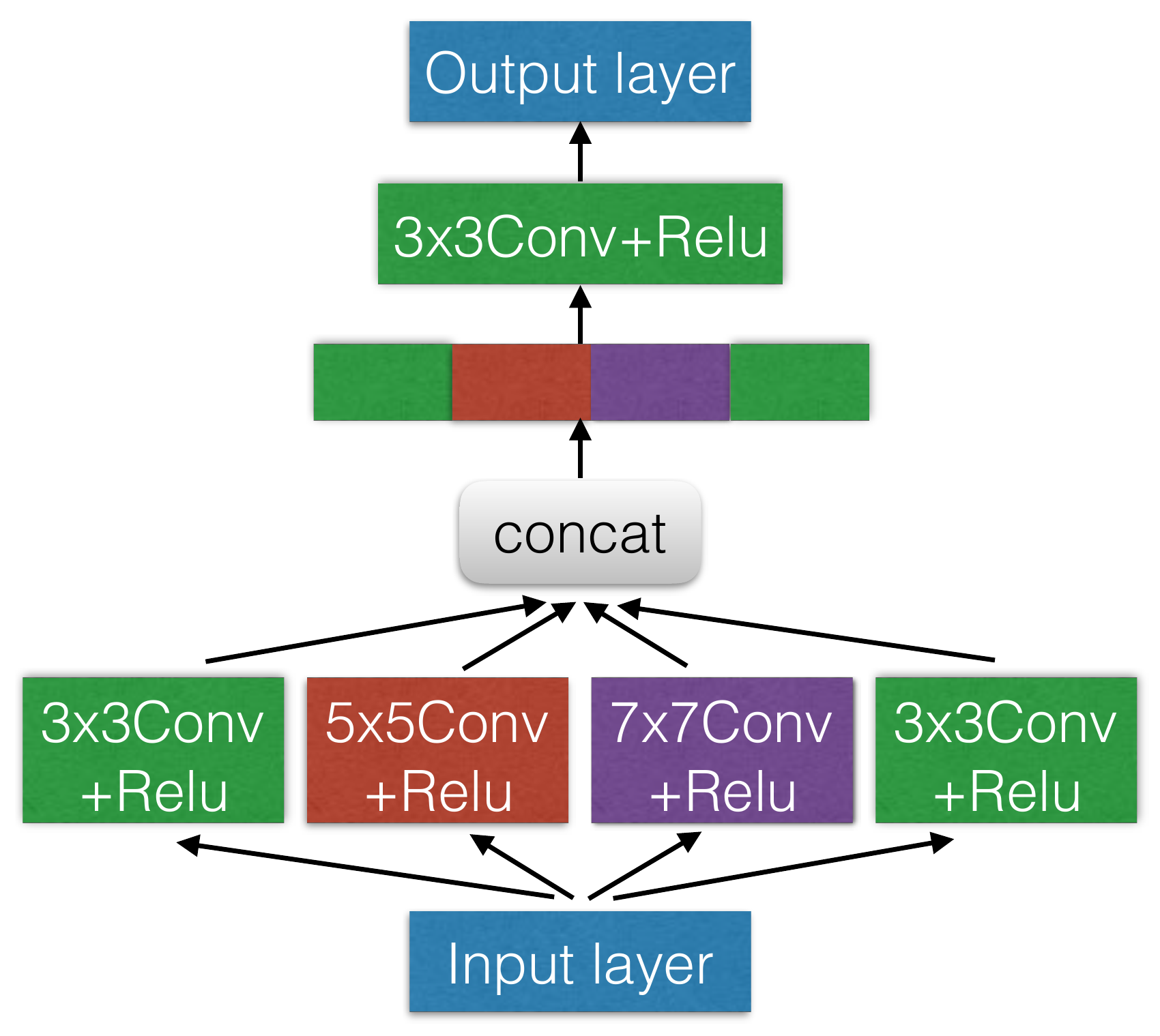}}
\subfigure[mix recurrent block]{
\label{fig2b} 
\includegraphics[width=4.5cm]{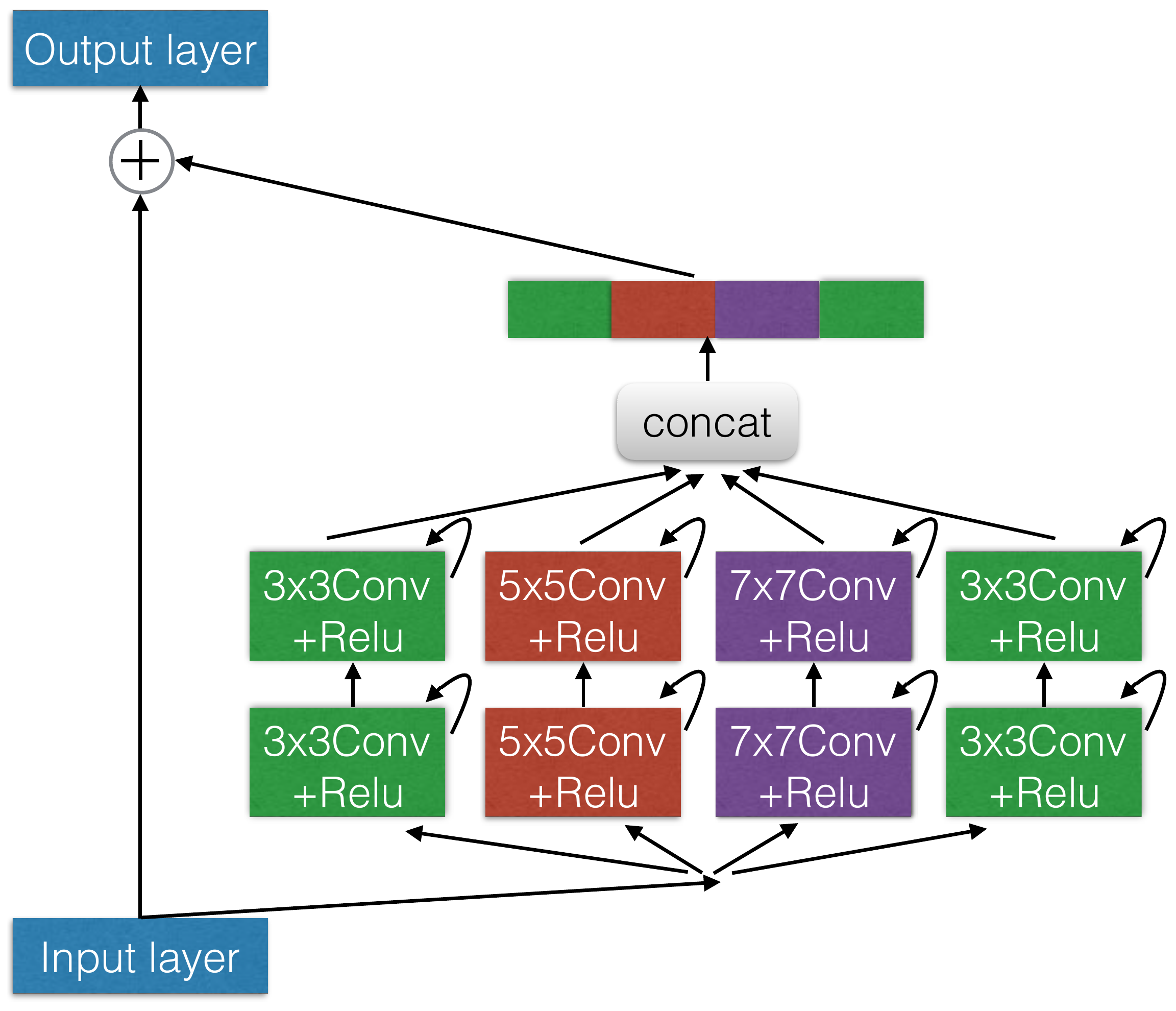}}
\caption{MixModule block in U-Net and its variants}
\label{fig2} 
\end{figure}

In this paper, we let n equals 4 and choose three convolution kernel sizes which are $3\times 3$, $5\times 5$ and $7\times 7$. Figure~\ref{fig1} and Figure~\ref{fig2} show the details of the modules. Figure~\ref{fig1} is the basic modules used in U-Nnet and its variants and Figure~\ref{fig2} is the corresponding MixModule version.

\begin{figure}[htb!]
\centering
\subfigure[U-Net/MixU-Net]{
\label{fig3a} 
\includegraphics[width=6.8cm]{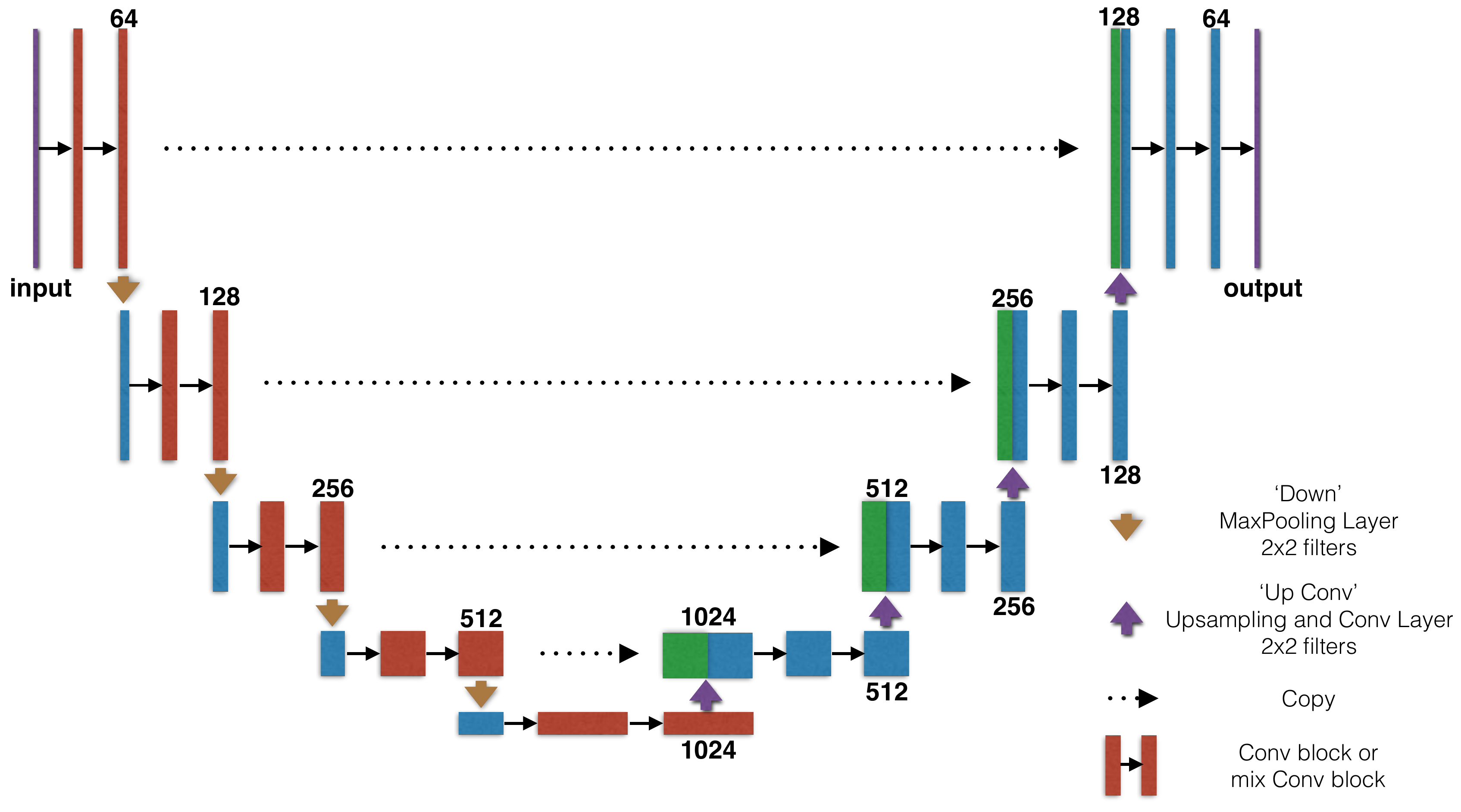}}
\subfigure[R2U-Net/MixR2U-Net]{
\label{fig3b} 
\includegraphics[width=6.8cm]{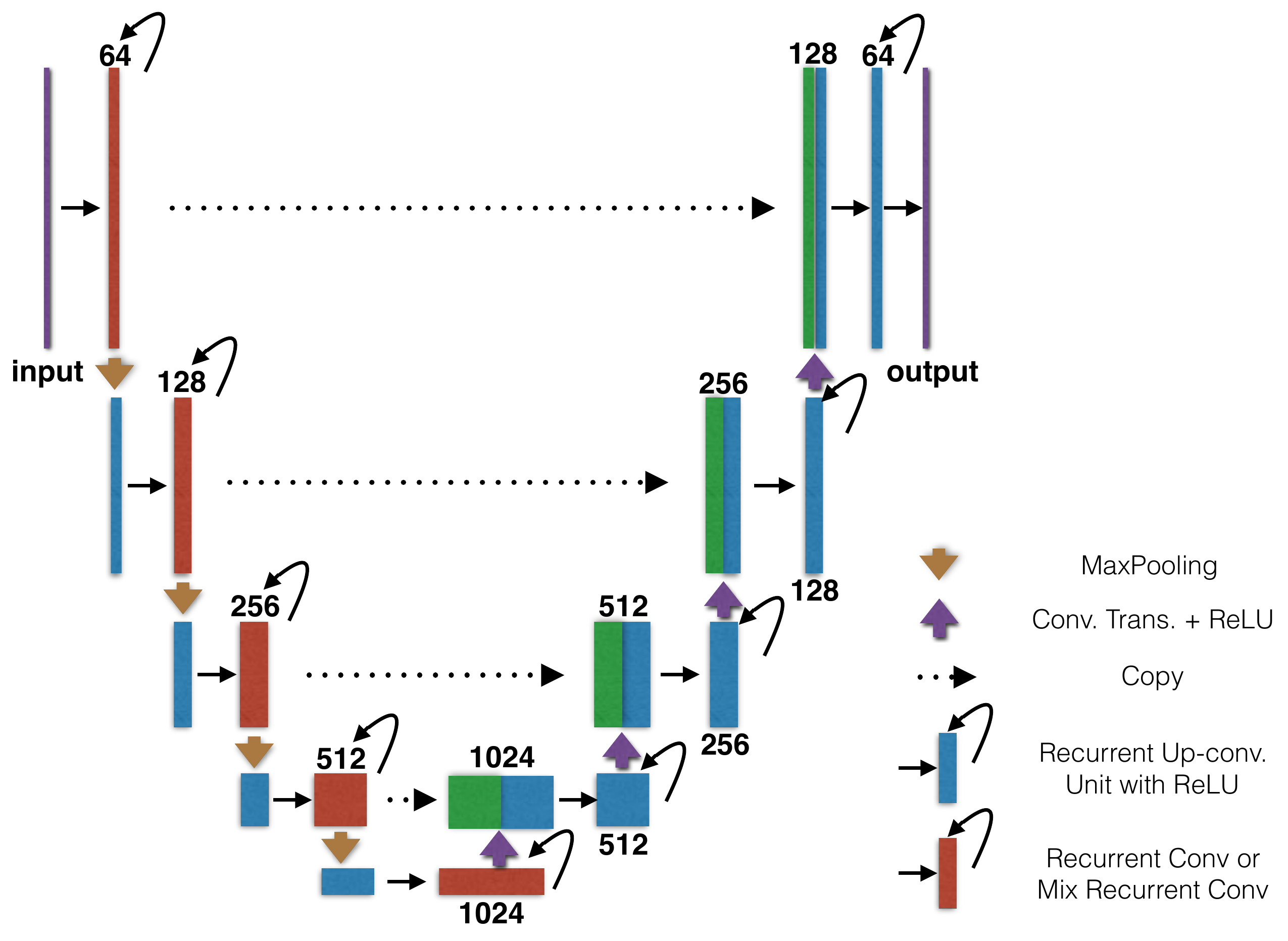}}
\subfigure[AttU-Net/MixAttU-Net]{
\label{fig3c} 
\includegraphics[width=6.8cm]{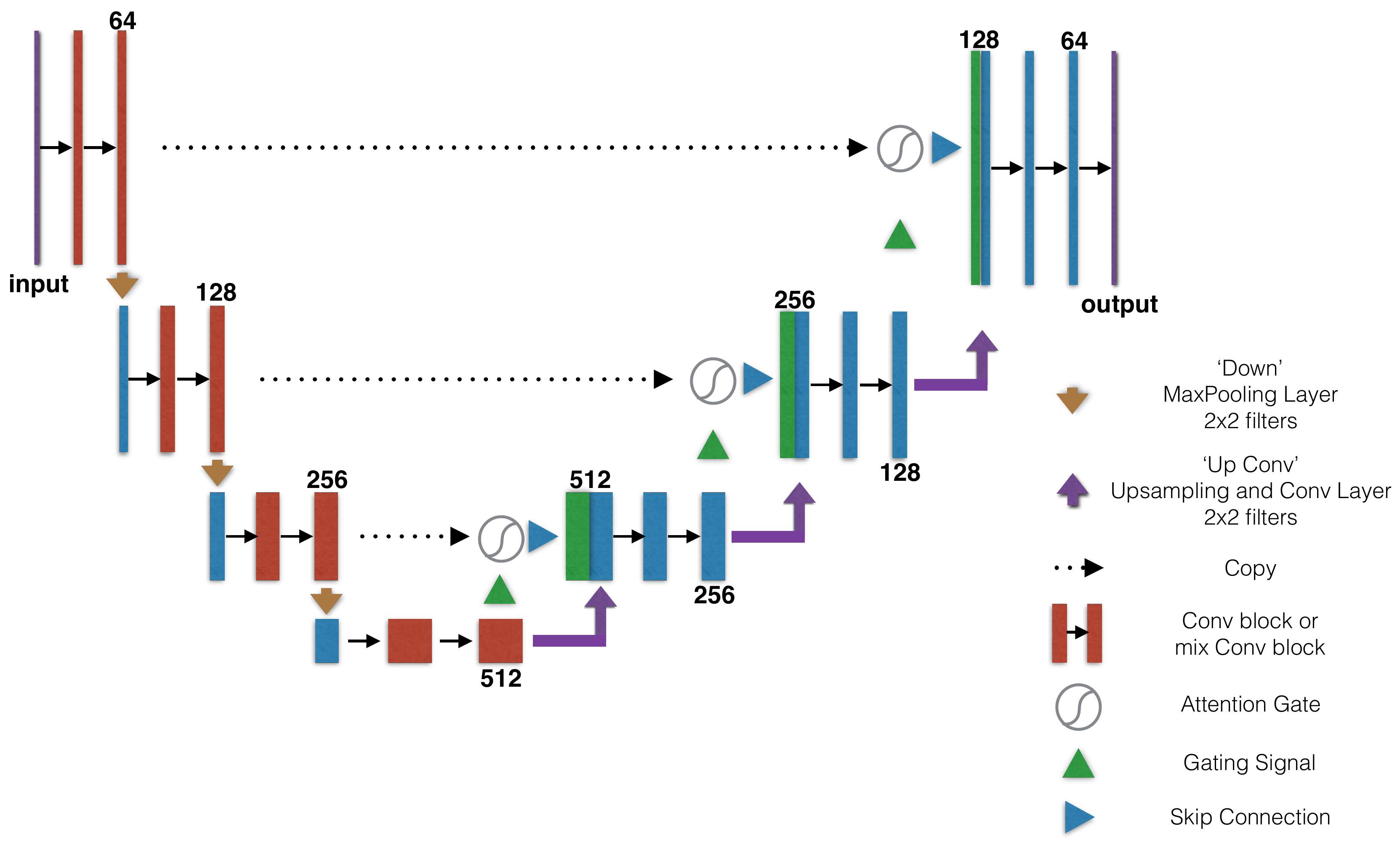}}
\caption{Basic Architecture and MixModule Architecture}
\label{fig3} 
\end{figure}

\subsection{Neural Network Architecture}
\label{ssec:arch}
We use MixModule to replace the single-size convolution kernel in the original U-Net and its variants. In this paper, we use U-Net\cite{ronneberger2015u}, R2U-Net\cite{alom2018recurrent} and Attention U-Net\cite{oktay2018attention}(AttU-Net for short) for experiments, whose network structures are shown in the Figure~\ref{fig3} and the brown module indicates the location of the replacement.

\section{EXPERIMENT AND RESULTS}
\label{sec:exp}
To demonstrate the effects of MixModule, we performe experiments on two different medical image datasets which include 2D images for skin lesion segmentation and retina blood vessel segmentation (DRIVE and CHASE\_DB1). We use PyTorch \cite{paszke2017automatic} framework to implement all the experiments on a single GPU machine with an NIVIDIA Quadro P6000.

\subsection{Dataset}
\subsubsection{Skin Lesion Segmentation}
\label{sssec:skin}
This dataset comes from ISIC Skin Image Analysis Workshop and Challenge of MICCAI 2018 \cite{codella2019skin} \cite{tschandl2018ham10000} and contains 2594 samples in total. The dataset was split into training set(70$\%$), validation set(10$\%$), and test set(20$\%$) which means 1815 images for training, 259 for validation and 520 for testing models. The original samples were slightly different in size from each other and were resized to 192$\times$256. 

\subsubsection{Retina Blood Vessel Segmentation}
\label{sssec:blood}
We perform retina blood vessel segmentation experiments on two different datasets, DRIVE\cite{staal2004ridge} and CHASE\_DB1\cite{fraz2012blood}. DRIVE dataset consists of 40 retinal images in total, in which 20 samples are used for training and remaining 20 for testing. The size of each original image is 565$\times$584 pixels and all images are cropped and padded with zeros to 576$\times$576 to get a square dataset. We randomly select 531265 patches whose size is 48$\times$48 from 20 of the training images in DRIVE dataset and 10$\%$ of them are used for validation. Another dataset, CHASE\_DB1, contains 28 color retina images with the size of 999$\times$960 pixels which are collected from both left and right eyes of 14 school children. 20 samples are randomly selected as training set and the remaining 8 samples are used for testing.
Similar to DRIVE dataset, we crop all the samples into 960$\times$960 pixels and randomly select 412400 patches of 48$\times$48 pixels from the training set of which 10$\%$ are used for validation and the remaining for training.

\subsection{Quantitative Analysis}
\label{ssec:qua}
To make a detailed comparison and analysis of the model performance, several quantitative analysis metrics are considered, including accuracy (AC)\eqref{con:ac}, sensitivity (SE)\eqref{con:se}, specificity (SP)\eqref{con:sp}, precision (PC)\eqref{con:pc}, Jaccard similarity (JS)\eqref{con:js} and F1-score (F1)\eqref{con:f1} which is also known as Dice coefficient (DC). Variables involved in these formulas are: True Positive (TP), True Negative (TN), False Positive (FP), False Negative (FN), Ground Truth(GT) and Segmentation Result (SR).
In the experiments, we utilize these metrics to evaluate the
performance of the proposed approaches against existing ones.
\begin{small}
\begin{equation}
AC=\frac{TP+TN}{TP+TN+FP+FN}\label{con:ac}
\end{equation}
\end{small}
\begin{small}
\begin{equation}
SE=\frac{TP}{TP+FN}\label{con:se}
\end{equation}
\end{small}
\begin{small}
\begin{equation}
SP=\frac{TN}{TN+FP}\label{con:sp}
\end{equation}
\end{small}
\begin{small}
\begin{equation}
PC=\frac{TP}{TP+FP}\label{con:pc}
\end{equation}
\end{small}
\begin{small}
\begin{equation}
JS=\frac{\left|GT\cap{SR}\right|}{\left|GT\cup{SR}\right|}\label{con:js}
\end{equation}
\end{small}
\begin{small}
\begin{equation}
F1=2 \frac{SE * PC}{SE + PC}, 
DC=2 \frac{\left|GT\cap{SR}\right|}{\left|GT\right|+\left|SR\right|}\label{con:f1} 
\end{equation}
\end{small}

\begin{figure}[htb!]
\centering
\subfigure[Skin]{
\label{fig_skin} 
\begin{minipage}{0.33\linewidth}
\includegraphics[width=2.66cm]{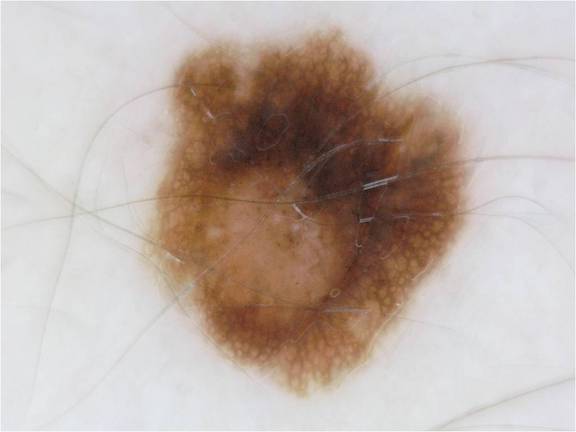}
\includegraphics[width=2.66cm]{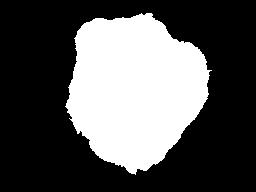}
\includegraphics[width=2.66cm]{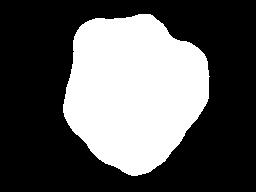}
\includegraphics[width=2.66cm]{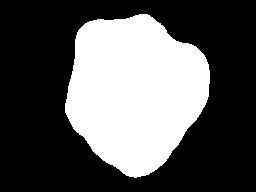}
\includegraphics[width=2.66cm]{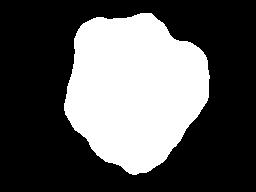}
\end{minipage}}%
\subfigure[DRIVE]{
\label{fig_drive} 
\begin{minipage}{0.25\linewidth}
\includegraphics[width=2cm]{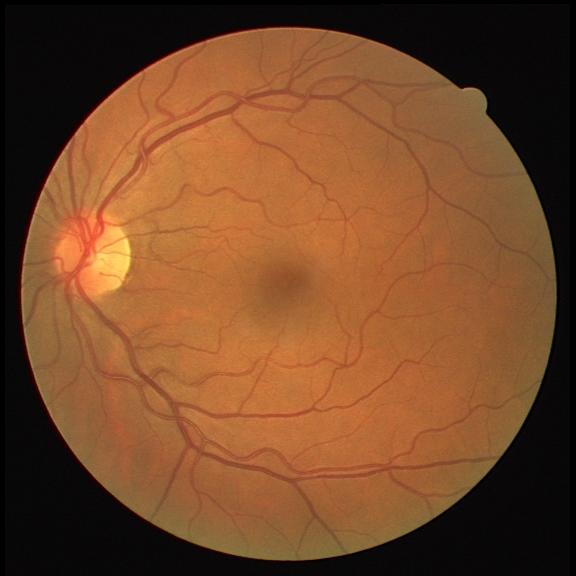}
\includegraphics[width=2cm]{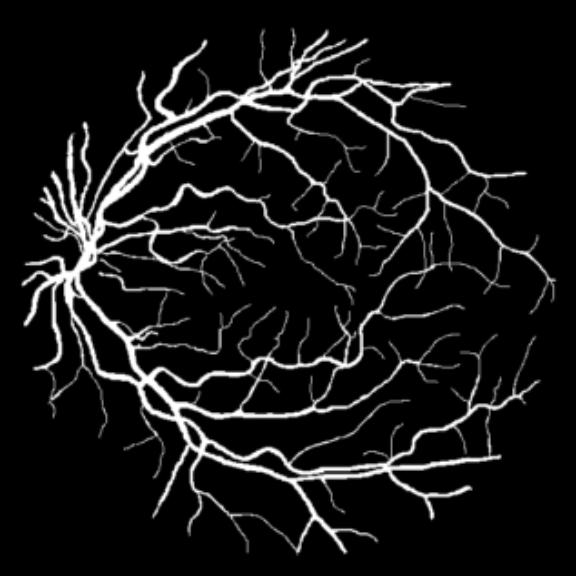}
\includegraphics[width=2cm]{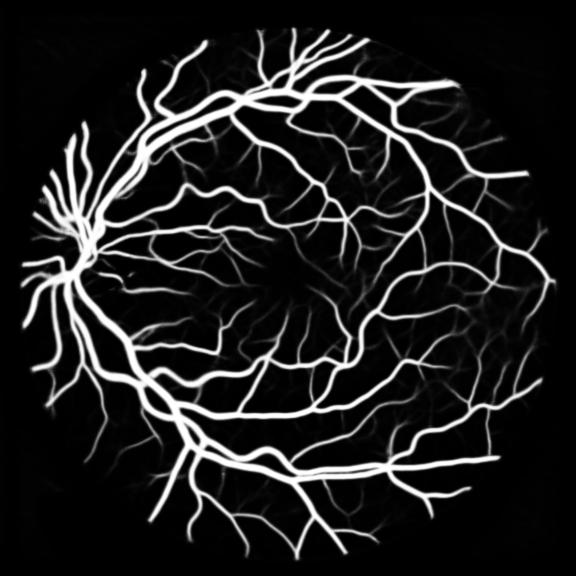}
\includegraphics[width=2cm]{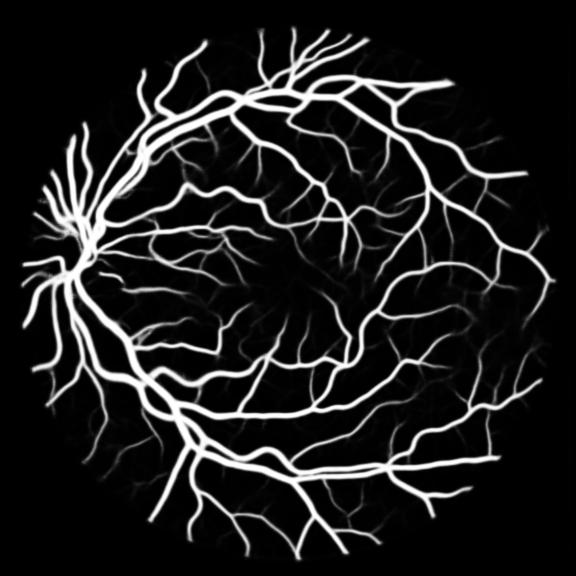}
\includegraphics[width=2cm]{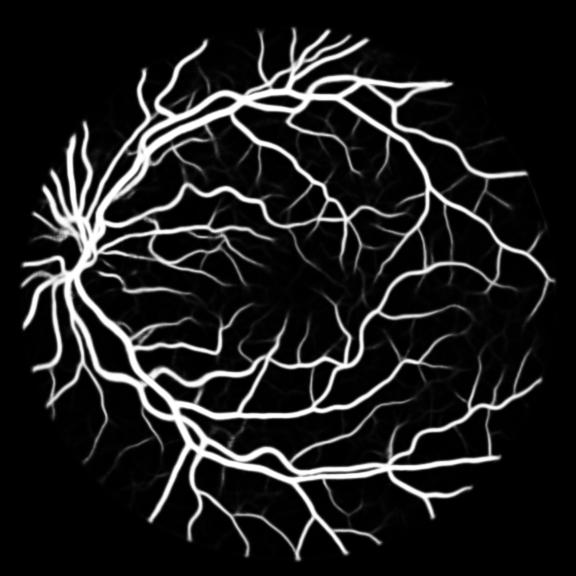}
\end{minipage}}%
\subfigure[CHASE\_DB1]{
\label{fig_chasedb1} 
\begin{minipage}{0.25\linewidth}
\includegraphics[width=2cm]{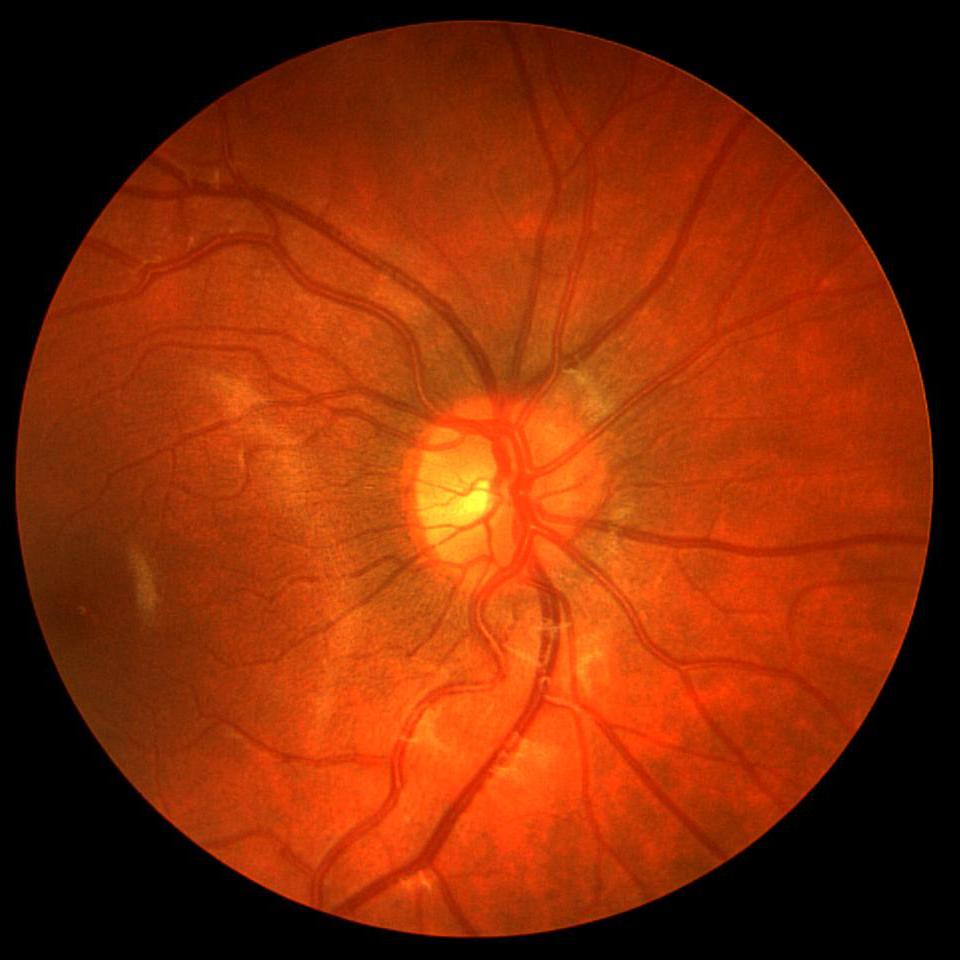}
\includegraphics[width=2cm]{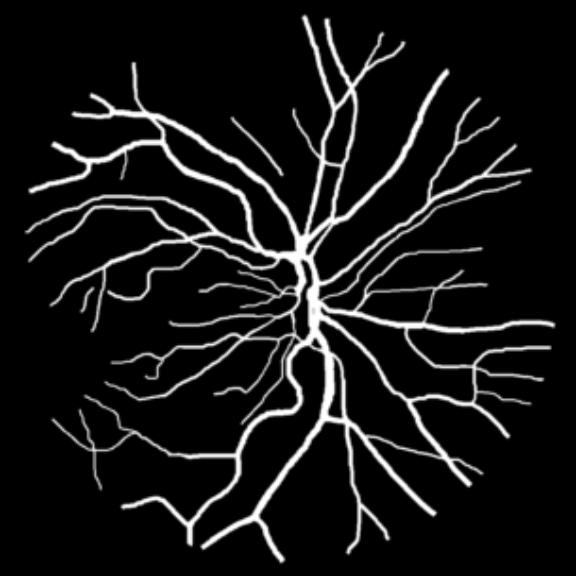}
\includegraphics[width=2cm]{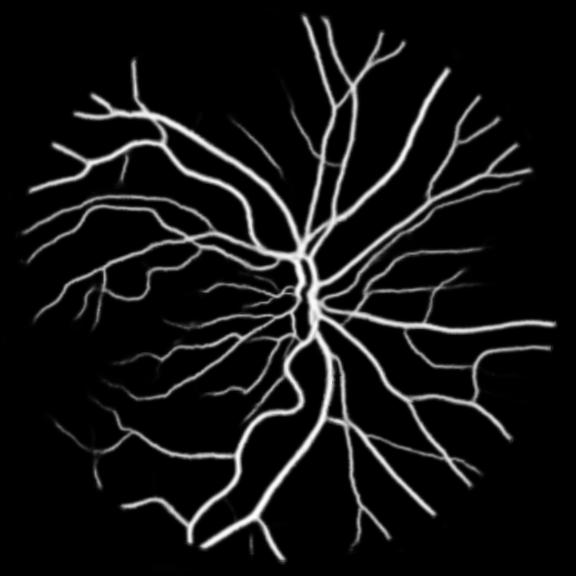}
\includegraphics[width=2cm]{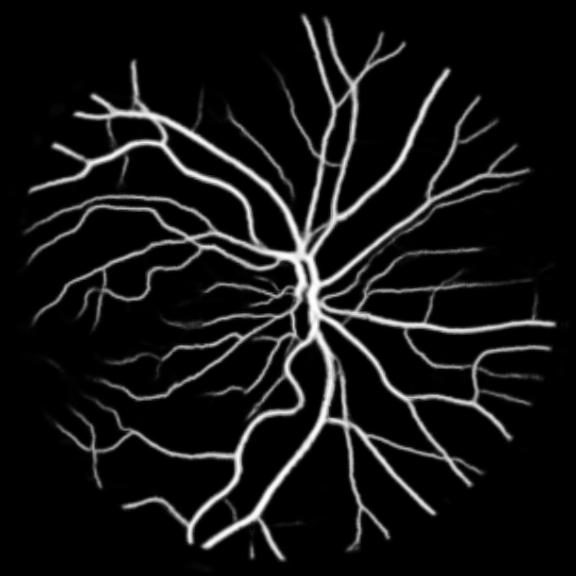}
\includegraphics[width=2cm]{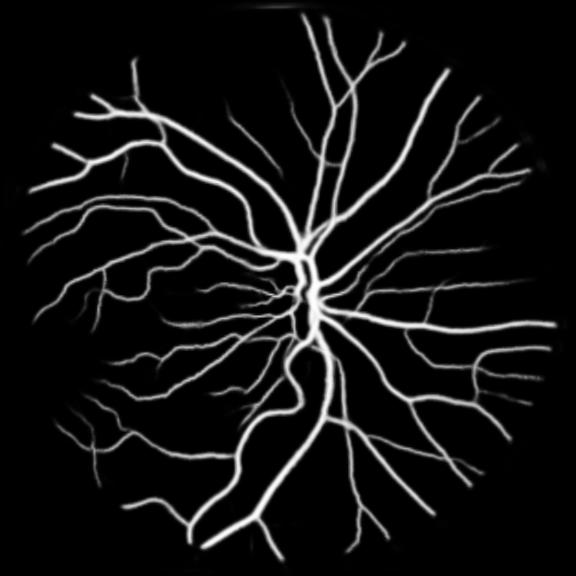}
\end{minipage}}%
\caption{Segmentation outputs of samples. From top to bottom: input images, ground truth, MixU-Net outputs, MixR2U-Net outputs and MixAttU-Net outputs}
\label{fig4} 
\end{figure}

\begin{table*}[h]
    \caption{Experimental Results on Three Datasets}
    \vspace{20pt}
    \centering
    \begin{tabular}{cccccccc}
        \hline
        Dateset& Methods& AC& SE& SP& PC& F1& JS\\
        \hline
        Skin  & U-Net & 0.9465 & 0.8154 & 0.9828 & 0.9218 & 0.8653 & 0.7645\\
        & MixU-Net & 0.9479 & 0.8294 & 0.9843 & \textbf{0.9312} & 0.8774 & 0.7673\\
        & R2U-Net & 0.9436 & 0.8111 & 0.9843 & 0.9126 & 0.8471 & 0.7446\\
        & MixR2U-Net & 0.9444 & 0.8116 & \textbf{0.9846} & 0.9248 & 0.8547 & 0.7573\\   
        & AttU-Net & 0.9496 & 0.8512 & 0.9728 & 0.8985 & 0.8742 & 0.7801\\
        & MixAttU-Net & \textbf{0.9512} & \textbf{0.8607} & 0.9777 & 0.9140 & \textbf{0.8865} & \textbf{0.7804}\\           
        \hline
        DRIVE  & U-Net & 0.9565 & 0.7542 & 0.9831 & 0.8559 & 0.8014 & 0.6694\\
        & Mix U-Net & 0.9581 & \textbf{0.7615} & 0.9840 & 0.8638 & 0.8090 & 0.6798\\
        & R2U-Net & 0.9539 & 0.7452 & 0.9740 & 0.8762 & 0.8054 & 0.6613\\
        & MixR2U-Net & 0.9547 & 0.7514 & \textbf{0.9867} & 0.8788 & 0.8101 & 0.6809\\   
        & AttU-Net & 0.9564 & 0.7295 & 0.9864 & 0.8767 & 0.7959 & 0.6617\\
        & MixAttU-Net & \textbf{0.9591} & 0.7517 & 0.9864 & \textbf{0.8801} & \textbf{0.8105} & \textbf{0.6820}\\ 
        \hline
        CHASE\_DB1  & U-Net & 0.9540 & 0.7819 & 0.9778 & 0.8130 & 0.7971 & 0.6732\\
        & Mix U-Net & \textbf{0.9544} & \textbf{0.8029} & 0.9746 & 0.8289 & \textbf{0.8156} & \textbf{0.6776}\\
        & R2U-Net & 0.9436 & 0.7559 & 0.9771 & 0.8130 & 0.7834 & 0.6378\\
        & MixR2U-Net & 0.9487 & 0.7565 & \textbf{0.9912} & \textbf{0.8483} & 0.7998 & 0.6590\\   
        & AttU-Net & 0.9507 & 0.7487 & 0.9781 & 0.8240 & 0.7840 & 0.6457\\
        & MixAttU-Net & 0.9533 & 0.7660 & 0.9788 & 0.8323 & 0.7972 & 0.6636\\         
        \hline       
    \end{tabular}
    \label{table1}
\end{table*}

\subsection{Results}
\label{ssec:res}
All three datasets are processed by subtracting the mean and normalizing according to the standard deviation. We use Adam optimizer, set the initial learning rate to 0.001 which is reduced by ten times if the training set loss does not drop during 10 consecutive epochs. We augment data using rotation, crop, flip, shift, change in contrast, brightness and hue. We set batch size to 4 for Skin Dataset and 32 for DRIVE and CHASE\_DB1 whose patch size is relatively smaller. For each model we train 50 epochs and the result is shown in Table~\ref{table1}. Models with MixModule have better performance than those not and the best performance in each metric all comes from MixModule-based models. We also show some outputs of the networks in Figure~\ref{fig4}.

\section{Conclusion}
\label{sec:con}
In this paper, we propose a new module named MixModule that can combine different ranges of features and can be embedded into different network structures of medical image segmentation tasks . We apply MixModule to U-Net and its two variants R2U-Net and Attention U-Net, get MixU-Net, MixR2U-Net and MixAttU-Net. These models are evaluated using three datasets including skin lesion segmentation and retina blood vessel segmentation. Experimental results show network models with MixModule has better performance than original ones in medical image segmentation tasks on all three datasets, which indicates MixModule has great development and application potential in medical image segmentation field.

\bibliographystyle{IEEEbib}
\bibliography{refs}

\end{document}